\documentclass[journal]{IEEEtran}

\usepackage{amsmath,amsfonts,amssymb}
\usepackage{algorithmic}
\usepackage{algorithm}
\usepackage{array}
\usepackage[caption=false,font=normalsize,labelfont=sf,textfont=sf]{subfig}
\usepackage{textcomp}
\usepackage{stfloats}
\usepackage{url}
\usepackage{verbatim}
\usepackage{graphicx}
\usepackage{cite}
\usepackage{hyperref}
\hypersetup{hypertex=true,
            colorlinks=true,
            linkcolor=blue,
            urlcolor=blue,
            citecolor=blue,
            }
\usepackage{booktabs}
\usepackage{multirow}
\usepackage{utfsym}
\usepackage{colortbl}
\usepackage{orcidlink}

\begin{document}

\title{ReIDMamba: Learning Discriminative Features with Visual State Space Model for Person Re-Identification}

\author{Hongyang Gu$^\dag$~\orcidlink{0000-0002-4987-5022}, Qisong Yang~\orcidlink{0000-0002-9686-2697}, Lei Pu, Siming Han and Yao Ding
        % <-this % stops a space
\thanks{This work was supported by the National Natural Science Foundation of China (No. 62401609), in part by the Natural Science Basis Research Plan in Shaanxi Province of China (No. 2024JC-YBQN-0628), in part by the China Postdoctoral Science Foundation (No. 2024M754275).}
\thanks{H. GU, Q. YANG, L. PU, S. HAN and Y. Ding are with the Rocket Force University of Engineering, Xi'an, China. (e-mail: guhy7@outlook.com, qisong.yang.93@outlook.com, warmstoner@163.com, hansm119@outlook.com, dingyao.88@outlook.com)}
\thanks{$^\dag$Corresponding author: guhy7@outlook.com}
\thanks{\copyright~2025 IEEE. Personal use of this material is permitted. Permission from IEEE must be obtained for all other uses, in any current or future media, including reprinting/republishing this material for advertising or promotional purposes, creating new collective works, for resale or redistribution to servers or lists, or reuse of any copyrighted component of this work in other works.}}

% The paper headers
\markboth{Journal of \LaTeX\ Class Files,~Vol.~14, No.~8, August~2021}%
{Shell \MakeLowercase{\textit{et al.}}: A Sample Article Using IEEEtran.cls for IEEE Journals}

\maketitle

\begin{abstract}
  Extracting robust discriminative features is a critical challenge in person re-identification (ReID).  While Transformer-based methods have successfully addressed some limitations of convolutional neural networks (CNNs), such as their local processing nature and information loss resulting from convolution and downsampling operations, they still face the scalability issue due to the quadratic increase in memory and computational requirements with the length of the input sequence. To overcome this, we propose a pure Mamba-based person ReID framework named ReIDMamba. Specifically, we have designed a Mamba-based strong baseline that effectively leverages fine-grained, discriminative global features by introducing multiple class tokens. To further enhance robust features learning within Mamba, we have carefully designed two novel techniques. First, the multi-granularity feature extractor (MGFE) module, designed with a multi-branch architecture and class token fusion, effectively forms multi-granularity features, enhancing both discrimination ability and fine-grained coverage. Second, the ranking-aware triplet regularization (RATR) is introduced to reduce redundancy in features from multiple branches, enhancing the diversity of multi-granularity features by incorporating both intra-class and inter-class diversity constraints, thus ensuring the robustness of person features. To our knowledge, this is the pioneering work that integrates a purely Mamba-driven approach into ReID research. Our proposed ReIDMamba model boasts only one-third the parameters of TransReID, along with lower GPU memory usage and faster inference throughput. Experimental results demonstrate ReIDMamba's superior and promising performance, achieving state-of-the-art performance on five person ReID benchmarks. Code is available at \href{https://github.com/GuHY777/ReIDMamba}{https://github.com/GuHY777/ReIDMamba}.

\end{abstract}


\section{Introduction}
\IEEEPARstart{P}{erson} re-identification (ReID) is a cornerstone task in the domain of computer vision, with significant applications in areas such as video surveillance and smart city infrastructure\cite{DBLP:journals/pami/YeSLXSH22}. Its primary goal is to extract discriminative features, enabling consistent person recognition across different camera views. With deep learning, feature extraction has shifted from traditional methods to sophisticated, data-driven approaches.
\begin{figure}[!t]
  \centering
  
  \includegraphics[width=0.5\textwidth]{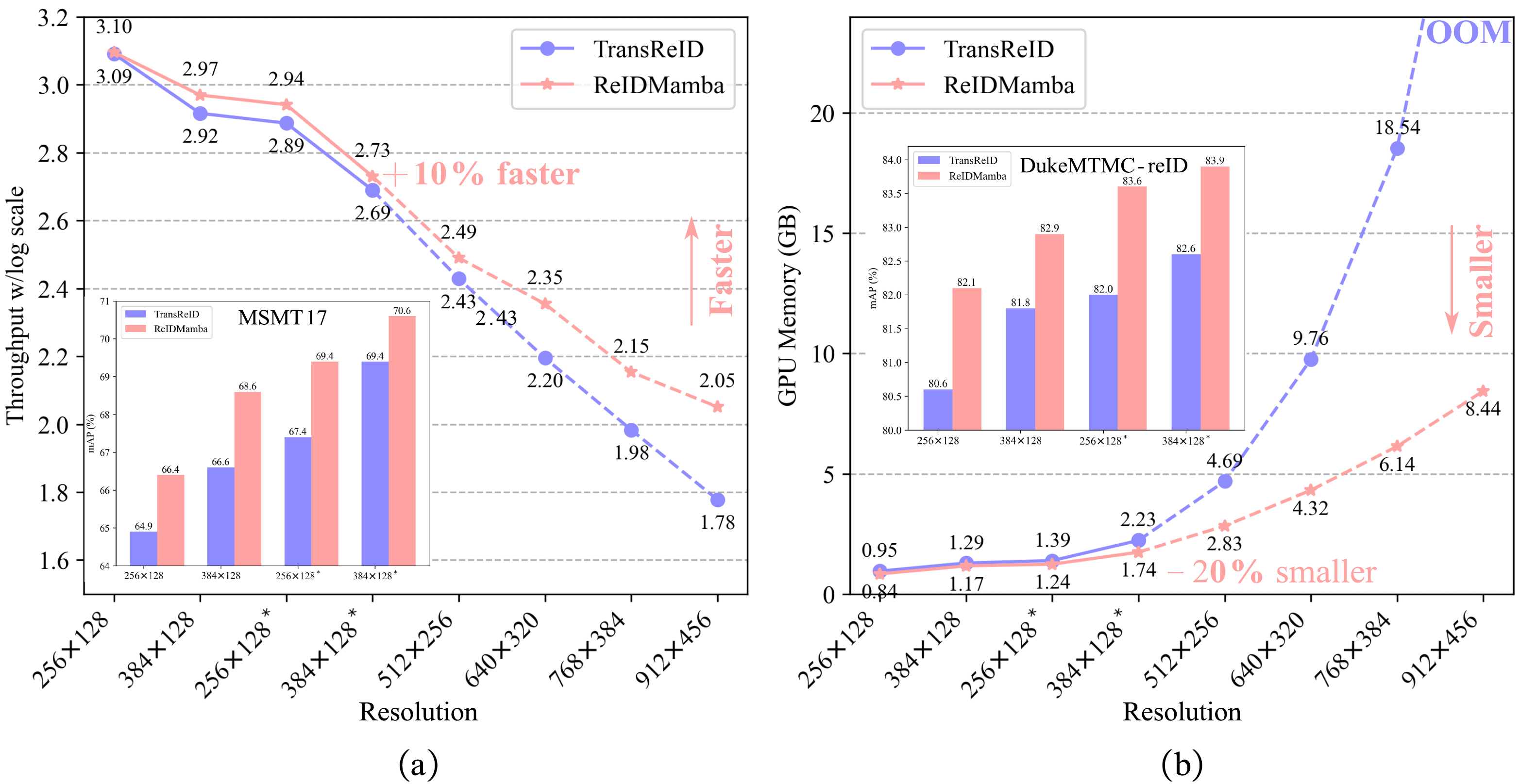}
  \caption{Performance and efficiency comparisons between TransReID and our ReIDMamba model regarding throughput (a) and GPU memory (b).}
  \label{fig:1}

\end{figure}

Convolutional Neural Networks (CNNs) have become the de facto standard for person ReID discriminative feature learning among deep learning methods, excelling at extracting hierarchical features from images to handle variations in appearance. Despite the significant progress achieved with CNNs\cite{DBLP:conf/mm/WangYCLZ18,DBLP:journals/tmm/LuoJGLLLG20,DBLP:journals/pami/ZhouYCX22}, their inherent focus on local features and fixed receptive fields often restricts global context capture, which is crucial for distinguishing persons with subtle differences. To address these limitations, researchers have integrated attention mechanisms into CNNs, which offers a more flexible way to improve their ability to capture global context and long-range dependencies\cite{DBLP:journals/pami/YeSLXSH22,DBLP:conf/cvpr/ZhangLZJ020,DBLP:conf/iccv/ChenDXYCYRW19}. Nevertheless, the inductive bias and downsampling within CNNs often lead to the loss of crucial detailed information, which can be harmful for person ReID tasks. Transformer-based architectures\cite{DBLP:journals/tiv/SarkerZU24} like TransReID\cite{DBLP:conf/iccv/He0WW0021} and AAformer\cite{zhu2023aaformer} mitigate downsampling and inductive bias, leveraging self-attention to simultaneously process the entire input sequence and capture complex, global, and fine-grained patterns. Although Transformers perform well in person ReID, their memory and computational demands grow quadratically with sequence length, which limits their applicability in real-world scenarios with constrained hardware.

Recently, the Mamba architecture\cite{DBLP:journals/corr/abs-2312-00752}, based on the state space model (SSM)\cite{kalman1960new}, has gained significant attention in natural language processing\cite{DBLP:journals/corr/abs-2408-15496, DBLP:journals/corr/abs-2404-16112} and computer vision\cite{DBLP:conf/icml/ZhuL0W0W24, DBLP:journals/corr/abs-2405-14858,DBLP:journals/corr/abs-2406-02395}. Mamba's performance is comparable to, and occasionally surpasses, that of Transformer models. Unlike Transformers, Mamba employs a selective state space mechanism that adaptively determines which information to propagate or discard, thereby accurately capturing global features. Furthermore, Mamba's computational complexity during  inference  scales nearly linearly with the number of tokens, rendering it highly scalable and well-suited for processing high-resolution images, especially in resource-constrained environments.

The Mamba architecture has  shown a unique capability to capture long-range, global features, while concurrently mitigating the high GPU memory consumption typical of Transformer-based models.
This dual advantage has significantly  motivated our exploration of a novel Mamba-based architecture tailored for the domain of person ReID.
We have named this innovative framework ReIDMamba.
Specifically, we have designed a Mamba-based strong baseline incorporating multiple class tokens, which  are designed to facilitate the extraction of fine-grained, discriminative global features. 
To further strengthen feature learning within Mamba, we introduce two innovative techniques.

First, we  develop the multi-granularity feature extractor (MGFE) module,  combining a multi-branch architecture with class tokens fusion. This integration  yields multi-granularity features that  enhance the model's discriminative power and provide a more detailed and nuanced feature representation. This is pivotal for accurate person ReID across varying scenarios and camera views.

Second, we incorporate ranking-aware triplet regularization (RATR) into our framework. This regularization technique  aims to reduce redundancy in features extracted by different branches of our architecture. RATR promotes feature diversity by incorporating both intra-class and inter-class diversity, thereby ensuring distinct features within each class and clear separation from other classes. This dual focus on diversity and separation is crucial for maintaining the robustness of person features across different camera views and under diverse conditions.

To the best of our knowledge, we are the first to  apply a pure Mamba-based architecture in the field of person ReID. The contributions of this paper are  summarized as follows: 
\begin{itemize}
  \item We propose a strong baseline utilizing a pure Mamba for person ReID tasks for the first time and achieves comparable performance with CNN-based and Transformer-based frameworks.
  \item We design a multi-granularity feature extractor (MGFE) module that integrates a multi-branch architecture with class tokens fusion to form multi-granularity features, which improves discrimination ability and provides more fine-grained coverage.
  \item We design a ranking-aware triplet regularization (RATR) technique to encourage diversity in features from multiple branches, ensuring the robustness of person features.
  \item The final framework ReIDMamba achieves state-of-the-art performance on five person ReID benchmarks including MSMT17, Market-1501, DukeMTMC-reID, CUHK03 and OccludedDuke.
\end{itemize}

\section{Related Work}
\subsection{Person ReID}
Person ReID has garnered increasing attention from researchers since its inception in 2006\cite{DBLP:conf/cvpr/GheissariSH06}, due to its significant practical and theoretical value. Research on person ReID has consistently focused on extracting discriminative features. With the advent of the deep learning era, most state-of-the-art methods have leveraged the architecture of CNNs. IDE\cite{DBLP:journals/corr/ZhengYH16}, which utilizes ResNet-50\cite{DBLP:conf/cvpr/HeZRS16} as its backbone network and employs cross-entropy loss for identity classification, has provided a CNN-based benchmark in person ReID. Recognizing the differences between person ReID and classification problems, TriNet\cite{DBLP:journals/corr/HermansBL17} adopts triplet loss to enhance the discriminability of features. However, subsequent studies\cite{DBLP:conf/mm/WangYCLZ18, DBLP:journals/tmm/LuoJGLLLG20, DBLP:conf/cvpr/ZhengDSJGYHJ19, DBLP:conf/mm/HeLLLCM23} have shown that using both cross-entropy loss and triplet loss can improve the discriminability of features while ensuring convergence speed.

To enhance the ability of CNNs to extract fine-grained features, many methods capture fine-grained cues from different parts or regions. The fine-grained parts are either automatically generated by roughly horizontal stripes or  semantic parsing. Methods such as PCB\cite{DBLP:conf/eccv/SunZYTW18}, MGN\cite{DBLP:conf/mm/WangYCLZ18}, and Pyramid\cite{DBLP:conf/cvpr/ZhengDSJGYHJ19} divide an image into several stripes and extract local features for each stripe. Using parsing or keypoint estimation to align different parts between two persons has also proven effective for occluded person ReID\cite{DBLP:journals/pami/HouMCGSC22, DBLP:conf/wacv/SomersVA23}. Additionally, a considerable amount of work has focused on introducing attention mechanisms to enhance the long-range feature extraction capabilities of CNNs, with representative methods  such as HA-CNN\cite{li2018harmonious}, RAG-SC\cite{DBLP:conf/cvpr/ZhangLZJ020}, and AGW\cite{DBLP:journals/pami/YeSLXSH22}.

However, the inductive bias and downsampling within CNNs often lead to the loss of crucial detailed information, which can be detrimental to person ReID tasks. The TransReID framework\cite{DBLP:conf/iccv/He0WW0021} introduces the pure attention architecture of Transformers into the person ReID domain, leveraging the bias-free and downsampling-free characteristics of Transformers to achieve excellent performance.  Following TransReID, numerous Transformer-based person ReID frameworks have emerged\cite{DBLP:conf/cvpr/ZhuKLLTS22, DBLP:journals/corr/abs-2408-16684}. Similarly, by  incorporating auxiliary information such as key points and human parsing, Transformer-based methods have pushed the performance of occluded person ReID to new heights,  exemplified by AAformer\cite{zhu2023aaformer}, FCFormer\cite{DBLP:journals/tmm/WangLLLBGL24}, and PFD\cite{DBLP:conf/iccv/MiaoWLD019}. However, the quadratic increase in memory and computational requirements as the sequence length grows  hinders the practical deployment of Transformer-based methods in real-world applications with  constrained hardware resources.

\subsection{Vision Mamba}
Among the attempts to improve scaling efficiency, a feature that Transformers lack, state space models (SSMs) have emerged as compelling alternatives, attracting significant attention from the research community. Among these models, a selective SSM block known as Mamba \cite{DBLP:journals/corr/abs-2312-00752} incorporates structured SSMs with hardware-aware state expansion, leading to a highly efficient recurrent architecture  competitive with Transformers. Building upon the Mamba block, a series of follow-up studies have investigated the application of SSMs in computer vision. Vision Mamba (Vim), a straightforward vision Mamba model,  outperforms Vision Transformers \cite{DBLP:conf/iclr/DosovitskiyB0WZ21} by sequentially stacking Mamba blocks. Furthermore, VMamba \cite{liu2024vmamba} features a core module called the 2D Selective Scan (SS2D) module, which traverses along four scanning routes. The SS2D module bridges the gap between the ordered nature of 1D selective scan and the nonsequential structure of 2D vision data, facilitating the gathering of contextual information from various sources and perspectives, and has achieved performance that surpasses Transformers on various visual tasks. Mamba$^\circledR$ \cite{DBLP:journals/corr/abs-2405-14858} overcomes the feature artifacts exhibited in Vim by introducing registers, successfully expanding the scaling capabilities of Vision Mamba. The study of Mamba-based architectures continues to  advance \cite{yue2024medmamba,liu2024lightweight}.

Currently, some works have applied SSMs to multi- or cross-modal person ReID, like MambaReID\cite{DBLP:journals/sensors/ZhangXYW24}, MambaPro\cite{DBLP:journals/corr/abs-2412-10707}, and ReMamba\cite{geng2024remamba}. Our ReIDMamba framework, however,  addresses single-modal person ReID. In terms of architecture, ReIDMamba  employs a pure Mamba structure, unlike ReMamba's CNN-Mamba hybrid approach. This makes ReIDMamba simpler and more efficient in speed and memory usage. MambaReID and MambaPro  are designed for multi-modal ReID. Specifically, MambaReID  employs a down-sampling layer\cite{liu2024vmamba}, causing crucial detailed-information loss\cite{DBLP:conf/iccv/He0WW0021}. MambaPro, similar to our approach, leverages SSMs for efficient feature learning but focuses only on multi-modal aggregation and synergistic prompt tuning. These task-focus and model-design differences highlight ReIDMamba's unique contributions.

\section{Methodology}
Our person ReID framework is rooted in Mamba-based image classification, with several pivotal enhancements to capture robust features, resulting in the Mamba-based strong baseline  detailed in Sec.\ref{sec:mbsb} and depicted in Fig.\ref{fig:2}. To further enhance robust feature learning within the Mamba framework, a multi-granularity feature extractor (MGFE) and a ranking-aware triplet regularization (RATR) are meticulously designed, as discussed in Sec.\ref{sec:fgfe} and Sec.\ref{sec:datr}, respectively. These contributions culminate in the development of ReIDMamba, illustrated in Fig.\ref{fig:3}.

\subsection{Mamba-based Strong Baseline}\label{sec:mbsb}
\begin{figure}[!t]
  \centering
  
  \includegraphics[width=0.5\textwidth]{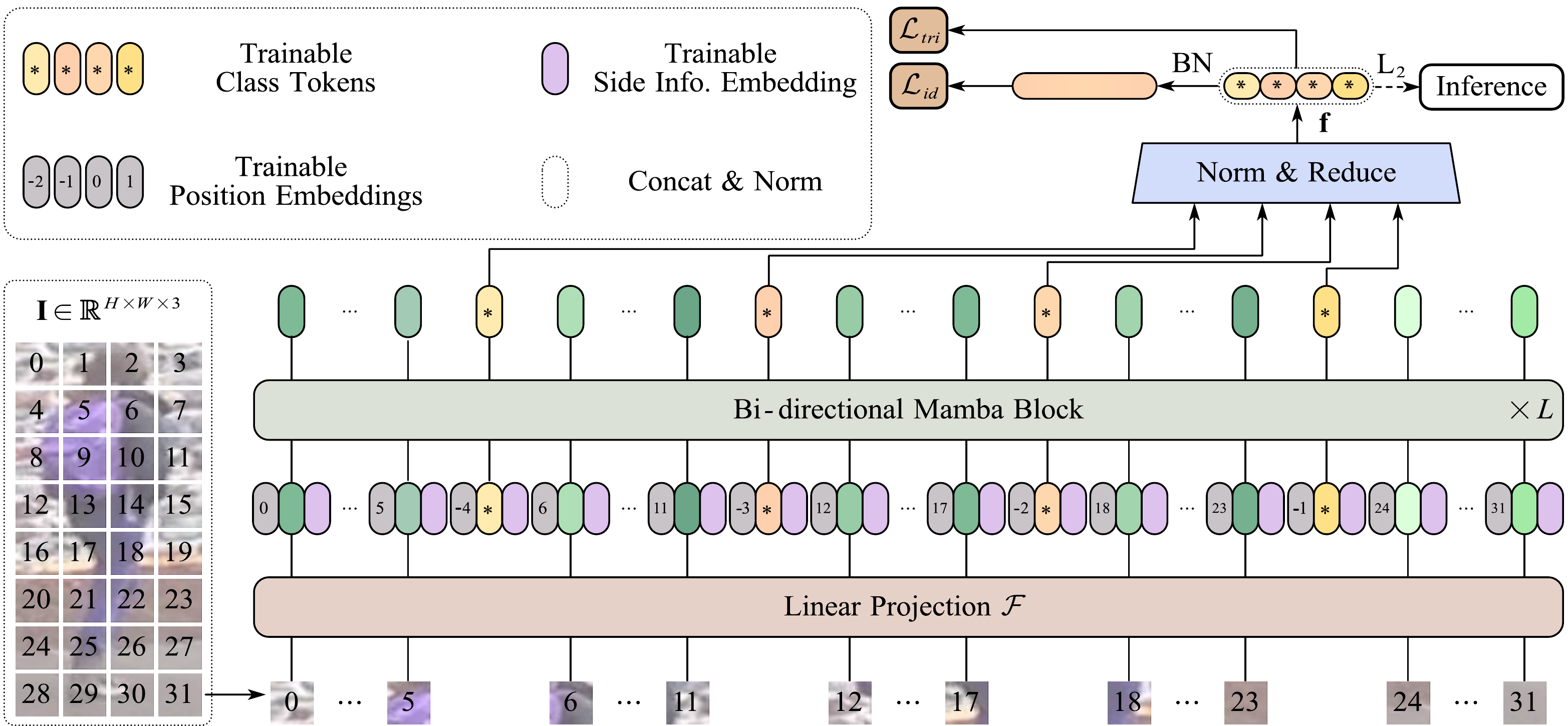}
  \caption{Mamba-based strong baseline framework (a nonoverlapping partition is shown, $M=4,N=32$). $M$ class tokens are evenly distributed among $N$ image tokens to extract fine-grained global person features. Subsequently, the $M$ class tokens are concatenated to form the final features used for ReID.
}
  \label{fig:2}
  
\end{figure}

We primarily develop a Mamba-based strong baseline for person ReID to extract robust features based on Mamba$^\circledR$\cite{DBLP:journals/corr/abs-2405-14858}, a variant of Vision Mamba (Vim)\cite{DBLP:conf/icml/ZhuL0W0W24}.
As shown in Fig.\ref{fig:2}.
Given an image $\mathbf{I}\in \mathbb{R} ^{H\times W\times 3}$, we first split the image into $N$ patches $\{\mathbf{x}_p^i\in\mathbb{R}^{P^\prime\times P^\prime\times 3}|i=0,1,\cdots,N-1\}$ of size $P^\prime \times P^\prime$ with stride $S$. Specifically, the number of patches $N$ can be computed as:
\begin{equation}
  N=h\times w=\lfloor \left( H+S-P^\prime \right) /S \rfloor \times \lfloor \left( W+S-P^\prime \right) /S \rfloor .
\end{equation}
Then, these image patches are transformed into $D$-dimensional image tokens through a linear projection $\mathcal{F}:\mathbb{R}^{P^\prime\times P^\prime\times 3}\rightarrow \mathbb{R}^D$, and $M$ class tokens $\{\mathbf{x}_{cls}^i\in\mathbb{R}^D|i=0,1,\cdots,M-1\}$ are evenly inserted among them to aggregate the discriminative information.
Following the setting of TransReID \cite{DBLP:conf/iccv/He0WW0021}, a learnable position embedding $\mathcal{P}\in\mathbb{R}^{(M+N)\times D}$ and a learnable side information embedding $\mathcal{S}\in\mathbb{R}^{N_c\times D}$ is subsequently added to retain positional information and incorporate side information respectively, which can be formulated as:
\begin{equation}
  \begin{aligned} \label{eq:gs2}
    \mathbf{z}_0=[&\mathcal{F} \left( \mathbf{x}_{p}^{0:J} \right) ;\mathbf{x}_{cls}^{0};\mathcal{F} \left( \mathbf{x}_{p}^{J:2J} \right) ;\mathbf{x}_{cls}^{1};\cdots ;\\
    &\mathcal{F} \left( \mathbf{x}_{p}^{\left( M-1 \right) J:MJ} \right) ;\mathbf{x}_{cls}^{M-1};\mathcal{F} \left( \mathbf{x}_{p}^{MJ:N-1} \right) ]\\
    &+\mathcal{P} +\lambda \mathcal{S} [c],
  \end{aligned}
\end{equation}
where $J=\lfloor N/(M+1) \rfloor$, $\mathcal{F}(\mathbf{x}_p^{i:j}):=[\mathcal{F}(\mathbf{x}_p^i);\cdots;\mathcal{F}(\mathbf{x}_p^{j-1})]$, $\lambda$ is a hyperparameter that controls the strength of the side information, and $c\in\{0,1,\cdots,N_c-1\}$ is the camera ID.
We set  $\lambda=3.0$ in the whole experiment like TransReID\cite{DBLP:conf/iccv/He0WW0021}.

\begin{figure*}[!t]
  \centering
  
  \includegraphics[width=1.0\textwidth]{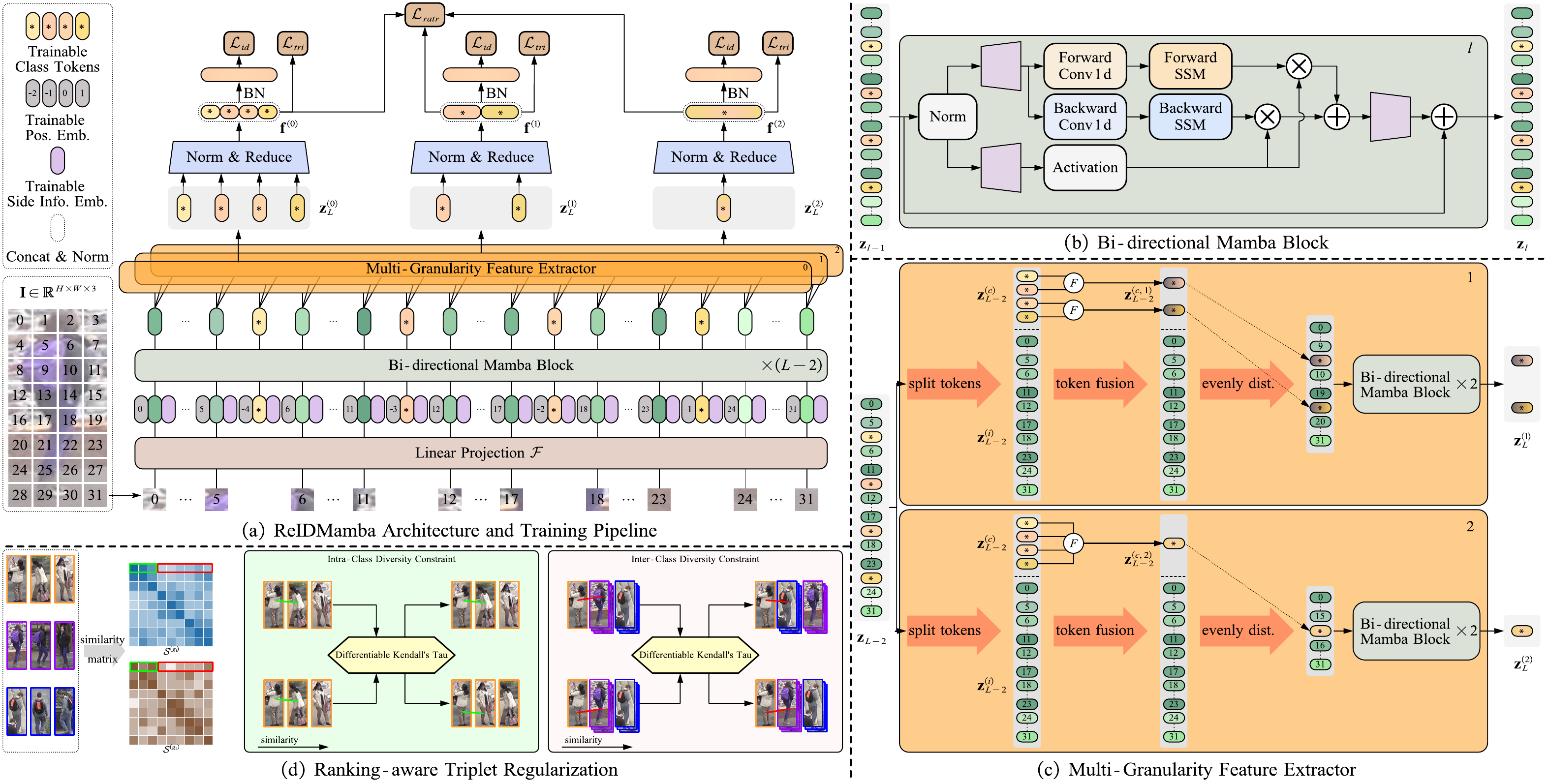}
  \caption{The overall architecture of ReIDMamba (a nonoverlapping partition is shown, $M=4,N=32,G=3$). (a) ReIDMamba, based on the Mamba-based strong baseline, incorporates a multi-branch architecture to extract multi-granularity features from multiple levels. The core block of ReIDMamba is the (b) bi-directional Mamba block within Vim. To ensure that each branch inherently maintains a multiple level of granularity, (c) multi-granularity feature extractor is used for tokens fusion. Finally, (d) ranking-aware triplet regularization is combined to enhance the diversity of features across various granularities.
  }
  \label{fig:3}
  
\end{figure*}

To map the input $\mathbf{z}_0$ into a latent representation space, we employ $L$ bi-directional Mamba blocks (BiMBs, as shown in Fig.\ref{fig:3}(b)), the core block in Vim.
For the $l$-th block, the input $\mathbf{z}_{l-1}$ is transformed into the output $\mathbf{z}_{l}$.
Inspired by the Mamba$^\circledR$, we implement a linear layer (LL) to reduce $M$ normalized output class tokens' dimensionality by a factor of $r$, and then concatenate them into a single vector in dimension of $M \times D/r$, which we refer to as the final global feature representation $\mathbf{f}$, as follows:
\begin{equation}
  \begin{aligned} \label{eq:ss3}
    \mathbf{z}_l&=\mathbf{BiMB}\left( \mathbf{z}_{l-1} \right) \in \mathbb{R} ^{\left( M+N \right) \times D},l=1,2,\cdots ,L\\
    \mathbf{z}_{L}^{\prime}&=\mathbf{Norm}\left( \mathbf{z}_{L}[J,2J+1,\cdots ,MJ+M-1] \right) \in \mathbb{R} ^{M\times D}\\
    \mathbf{z}_{L}^{\prime\prime}&=\mathbf{LL}\left( \mathbf{z}_{L}^{\prime} \right) \in \mathbb{R} ^{M\times (D/r)}\\
    \mathbf{f}&=\mathbf{Norm}\left( \left[ \mathbf{z}_{L}^{\prime\prime}\left[ 0 \right] ;\mathbf{z}_{L}^{\prime\prime}\left[ 1 \right] ;\cdots ;\mathbf{z}_{L}^{\prime\prime}\left[ M-1 \right] \right] \right) \in \mathbb{R} ^{MD/r}\\
  \end{aligned}.
\end{equation}

Inspired by BoT\cite{DBLP:journals/tmm/LuoJGLLLG20}, we introduce BNNeck to optimize the network. Specifically, the triplet loss $\mathcal{L}_{tri}$\cite{hermans2017defense} is directly applied to the global feature $\mathbf{f}$, while the ID loss $\mathcal{L}_{id}$\cite{DBLP:journals/tomccap/ZhengZY18} is applied to the global feature after batch normalization. Unlike TransReID\cite{DBLP:conf/iccv/He0WW0021}, the ID loss utilizes cross-entropy loss with label smoothing. Additionally, the triplet loss is adopted as a hard-margin approach rather than a soft-margin (The corresponding ablation studies can be found in Table~\ref{tab:ablations1}). Finally, the L2 normalized global feature $\mathbf{f}$ is utilized for inference.

\subsection{Multi-Granularity Feature Extractor} \label{sec:fgfe}
Although the Mamba-based strong baseline achieves impressive performance in person ReID, it only utilizes information from a single level, lacking the exploration of multi-granularity features. Inspired by MGN\cite{DBLP:conf/mm/WangYCLZ18}, we propose a Multi-Granularity Feature Extractor (MGFE) that integrates a multi-branch architecture to extract person features from various levels, as shown in Fig.\ref{fig:3}(c). Assuming there are a total of $G$ branches, denoted as $\{\textbf{MGFE}^{(g)} | g=0,1,\cdots,G-1\}$, each of them extracts features $\{\mathbf{z}_{L}^{(g)} | g=0,1,\cdots,G-1\}$ with multiple granularities based on $\mathbf{z}_{L-2} \in \mathbb{R}^{(M+N) \times D}$, which can be formulated as:
\begin{equation}
  \mathbf{z}_{L}^{(g)} = \mathbf{MGFE}^{(g)}\left( \mathbf{z}_{L-2} \right) \in \mathbb{R}^{\left( {M}/{2^g} \right) \times D}.
\end{equation}

Specifically, for branch $g$, we first split $\mathbf{z}_{L-2}$ into class tokens $\mathbf{z}_{L-2}^{(c)} \in \mathbb{R}^{M \times D}$ and image tokens $\mathbf{z}_{L-2}^{(i)} \in \mathbb{R}^{N \times D}$:
\begin{equation}
\left\{
\begin{aligned}
\mathbf{z}_{L-2}^{(c)} &:= \mathbf{z}_{L-2}[J, 2J+1, \cdots, MJ+M-1] \\
\mathbf{z}_{L-2}^{(i)} &:= \mathbf{z}_{L-2}[:J, J+1:2J+1, \cdots, MJ+M:]
\end{aligned}
\right. .
\end{equation}
Then, based on the fusion sampling rate $2^g$, we merge the class tokens $\mathbf{z}_{L-2}^{(c)} \in \mathbb{R}^{M \times D}$ by combining adjacent $2^g$ tokens to form $\mathbf{z}_{L-2}^{(c,g)}\in \mathbb{R}^{(M/2^g) \times D}$:
\begin{equation}
\mathbf{z}_{L-2}^{(c,g)}[j] = F\left( \mathbf{z}_{L-2}^{(c)}[j\times 2^g:(j+1)\times 2^g] \right),
\end{equation}
where $j=0,1,\cdots,M/2^g-1$, and $F$ can employ various fusion strategies, such as taking the average, maximum, or minimum. According to subsequent experiments, using the maximum operation yields the best results (as shown in Fig.\ref{fig:5}). 
The fused class tokens $\mathbf{z}_{L-2}^{(c,g)}$ and the previous image tokens $\mathbf{z}_{L-2}^{(i)}$ are mixed again according to the principle of evenly distribution, similar to Eq. \ref{eq:gs2}, to obtain $\mathbf{z}_{L-2}^{(g)}$, which will then undergo processing through 2 additional BiMBs to yield the final $\mathbf{z}_{L}^{(g)}$.

Through MGFE, we can obtain $G$ distinct fine-grained features $\{\mathbf{z}_L^{(g)}|g=0,1,\cdots,G-1\}$ (as shown in Fig. \ref{fig:8}). Similar to the Mamba-based strong baseline, each feature $\mathbf{z}_L^{(g)}$ is subjected to normalization, dimensionality reduction, and concatenation. To ensure that each fine-grained feature contributes equally, the concatenated feature dimension is set to be consistent. Specifically, for branch $g$, the dimensionality reduction factor is $r^{(g)}=r/2^g$, and the final dimension of $\mathbf{f}^{(g)}$ remains $MD/r$, consistent with the dimension of the feature $\mathbf{f}$ obtained from Eq.\ref{eq:ss3}.
For each $\mathbf{f}^{(g)}$, similar to the Mamba-based strong baseline, BNNeck is introduced for network optimization, which will be detailed in Sec.\ref{sec:trainingandinference}.

\subsection{Ranking-aware Triplet Regularization}\label{sec:datr}
\begin{figure}[!t]
  \centering
  
  \includegraphics[width=0.5\textwidth]{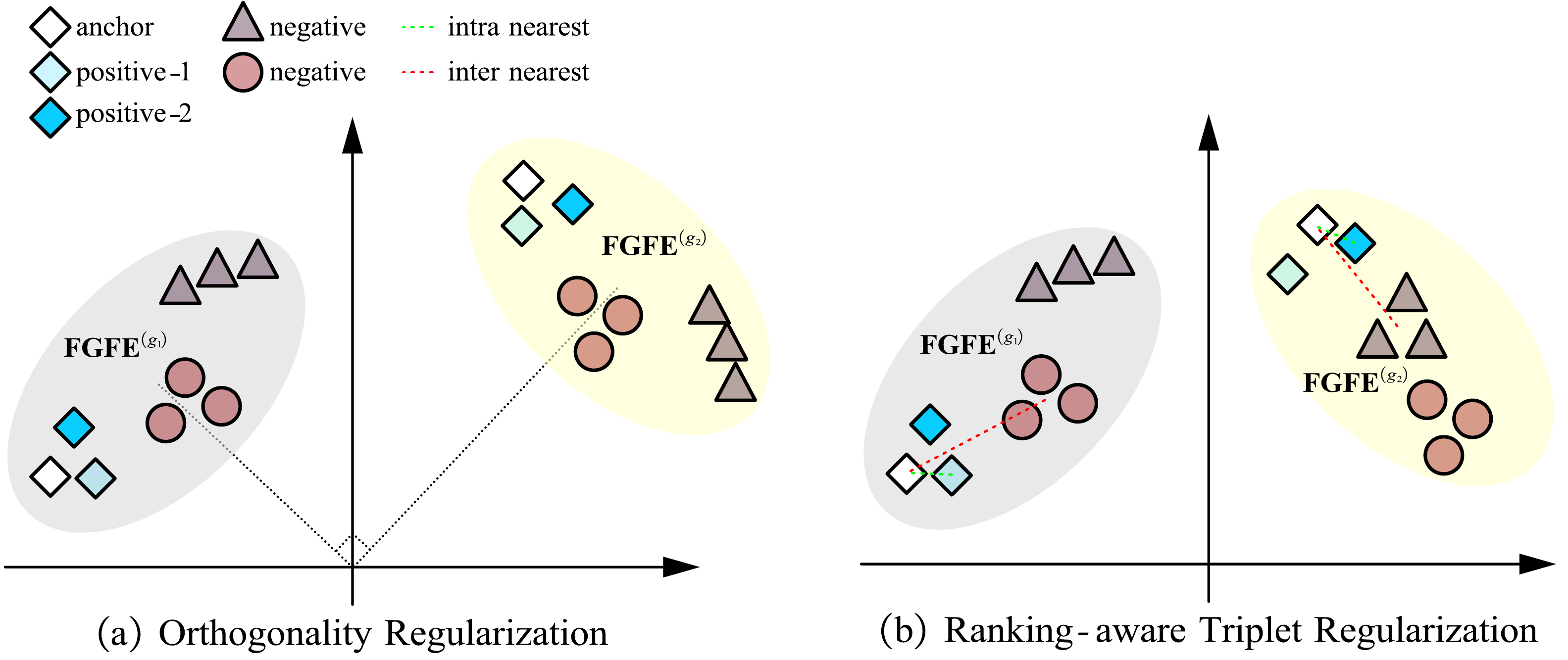}
  \caption{Comparison of different regularizations. (a) The features from the two branches are completely orthogonal, yet their relative similarity is identical, failing to achieve the goal of increasing diversity. (b) Ranking-aware triplet regularization enhances the diversity of the two features from both intra-class and inter-class perspectives.
  }
  \label{fig:4}
  
\end{figure}

Although we introduce multi-granularity features in $G$ branches with MGFE, the lack of explicit supervision makes it difficult to ensure the diversity among the features $\{\mathbf{f}^{(g)}|g=0,1,\cdots,G-1\}$ obtained by the $G$ branches, thereby limiting the robustness of the overall features. There are many regularization methods to enhance feature diversity, among which ABDNet\cite{DBLP:conf/iccv/ChenDXYCYRW19} enhance feature diversity by imposing orthogonal regularization on features. Considering the strong fitting ability of neural networks, the orthogonal regularization might allow features from different branches to maintain relative distances without genuinely promoting diversity, as shown in Fig.\ref{fig:4}(a). We propose ranking-aware triplet regularization (RATR) from the rankings of a minibatch of features, enhancing the diversity of features obtained by each branch from both intra-class and inter-class features perspectives, as shown in Fig.\ref{fig:4}(b).

Specifically, given a minibatch of image data with multi-granularity features $\{\mathbf{f}_i^{(g)}|g=0,1,\cdots,G-1; i=0,1,\cdots,PK-1\}$, where $P$ and $K$ denote the number of distinct identities of persons in the minibatch and the number of images of each identity, respectively. First, we compute the similarity matrix $\mathcal{S}^{(g)} \in \mathbb{R}^{PK \times PK}$ for the features of branch $g$:
\begin{equation}
  \mathcal{S} ^{\left( g \right)}[i,j]=\angle \left( \mathbf{f}_{i}^{\left( g \right)},\mathbf{f}_{j}^{\left( g \right)} \right),
\end{equation}
where $\angle(\cdot,\cdot)$ denotes the cosine similarity between two vectors. Then, we incorporate intra-class and inter-class diversity, expecting the rankings of features obtained by different branches to be distinct in terms of both intra-class and inter-class rankings. Kendall's Tau (KTau) is an indicator measuring the ranking correlation between two sequences. For sequences $\mathbf{x}\in\mathbb{R}^B$ and $\mathbf{y}\in\mathbb{R}^B$, it can be represented as: 
\begin{equation} 
  \text{KTau}(\mathbf{x}, \mathbf{y})=\frac{1}{C_{B}^{2}}\sum_{0 \leqslant i<j \leqslant B-1}{\text{sign}\left( \mathbf{x}_i-\mathbf{x}_j \right)}\cdot \text{sign}\left( \mathbf{y}_i-\mathbf{y}_j \right) ,
\end{equation} 
where $C_B^2$ is the combination, and $\text{sign}(\cdot)$ is the sign function. 
The range of KTau is $[-1, 1]$, where a smaller value indicates a less consistency in the ranking order between the two sequences. Therefore, to encourage diversity in the features learned by every branch, our goal is to minimize the KTau between the similarity matrices of the features learned by each pair of branches. We approach this from two perspectives: intra-class diversity and inter-class diversity, and propose the following intra-class diversity constraint loss and inter-class diversity constraint loss: 

\begin{footnotesize}
\begin{subequations}\label{eq:intra_inter_ori}
  \begin{align}
    \mathcal{L} _{ratr}^{(intra)}=\frac{1}{PK}\sum_{k=0}^{PK-1}{\sum_{0\leqslant i<j<G}{\frac{\text{KTau}\left( \mathcal{S} ^{\left( i \right)}\left[ k,\mathcal{K} _+ \right] ,\mathcal{S} ^{\left( j \right)}\left[ k,\mathcal{K} _+ \right] \right)}{C_{G}^{2}}}}
\\
\mathcal{L} _{ratr}^{(inter)}=\frac{1}{PK}\sum_{k=0}^{PK-1}{\sum_{0\leqslant i<j<G}{\frac{\text{KTau}\left( \mathcal{S} ^{\left( i \right)}\left[ k,\mathcal{K} _- \right] ,\mathcal{S} ^{\left( j \right)}\left[ k,\mathcal{K} _- \right] \right)}{C_{G}^{2}}}}
  \end{align}
\end{subequations}
\end{footnotesize}

\noindent where $\mathcal{K}_+$ denotes the set of positive indices of the $k$-th feature excluding $k$ itself, and $\mathcal{K}_-$ denotes the set of negative indices of the $k$-th feature, with $|\mathcal{K}_+| = K - 1$ and $|\mathcal{K}_-| = (P - 1)K$. However, due to the presence of the sign function, KTau is non-differentiable and thus Eq.\ref{eq:intra_inter_ori} cannot be directly optimized. To address this, we adopt the same approach as AutoLoss-GMS\cite{DBLP:conf/cvpr/Gu0FWC022} by replacing the sign function with the tanh function, thereby obtaining a differentiable KTau:

\begin{footnotesize}
  \begin{equation}\label{eq:dktau}
    \text{D-KTau}(\mathbf{x},\mathbf{y},\tau )=\frac{1}{C_{B}^{2}}\sum_{0\leqslant i<j\leqslant B-1}{\tanh \left( \frac{\mathbf{x}_i-\mathbf{x}_j}{\tau} \right) \cdot \tanh \left( \frac{\mathbf{y}_i-\mathbf{y}_j}{\tau} \right)},
  \end{equation}
\end{footnotesize}

\noindent where $\tau$ is a hyperparameter that controls the smoothness of the ranking correlation. Considering that each feature is associated with $(P - 1)K$ negative features, we calculate KTau based solely on the similarity between each feature and the centroid of the negative class features, thereby reducing the computation for each feature to only $P - 1$ negative features. We define the sequence as follows:
\begin{equation}
  \begin{aligned}
    \bar{\mathcal{S}}^{\left( g \right)}\left[ k,\mathcal{K} _- \right] =\left[ \angle \left( \mathbf{f}_{k}^{\left( g \right)},\bar{\mathbf{f}}_{c_1}^{\left( g \right)} \right) ,\angle \left( \mathbf{f}_{k}^{\left( g \right)},\bar{\mathbf{f}}_{c_2}^{\left( g \right)} \right) ,\cdots , \right.\\
    \left. \angle \left( \mathbf{f}_{k}^{\left( g \right)},\bar{\mathbf{f}}_{c_{P-1}}^{\left( g \right)} \right) \right]\\
  \end{aligned}
\end{equation}
where $\bar{\mathbf{f}}_{c_i}^{(g)}$ denotes the mean of all features within the negative class $c_i$ for the $k$-th feature in the minibatch for branch $g$. By incorporating the differentiable KTau, we can transform Eq.\ref{eq:intra_inter_ori} into ultimately optimizable and efficient loss functions:

\begin{footnotesize}
  \begin{subequations}\label{eq:intra_inter_new}
    \begin{align}
      \hat{\mathcal{L}}_{ratr}^{(intra)}=\frac{1}{PK}\sum_{k=0}^{PK-1}{\sum_{0\leqslant i<j<G}{\frac{\text{D-KTau}\left( \mathcal{S} ^{\left( i \right)}\left[ k,\mathcal{K} _+ \right] ,\mathcal{S} ^{\left( j \right)}\left[ k,\mathcal{K} _+ \right] ,\tau \right)}{C_{G}^{2}}}}
\\
\hat{\mathcal{L}}_{ratr}^{(inter)}=\frac{1}{PK}\sum_{k=0}^{PK-1}{\sum_{0\leqslant i<j<G}{\frac{\text{D-KTau}\left( \bar{\mathcal{S}}^{\left( i \right)}\left[ k,\mathcal{K} _- \right] ,\bar{\mathcal{S}}^{\left( j \right)}\left[ k,\mathcal{K} _- \right] ,\tau \right)}{C_{G}^{2}}}}
    \end{align}
  \end{subequations}
\end{footnotesize}

In summary, the proposed ranking-aware triplet regularization loss can be formulated as:
\begin{equation}
  \mathcal{L}_{ratr} = \hat{\mathcal{L}}_{ratr}^{(intra)} + \hat{\mathcal{L}}_{ratr}^{(inter)}.
\end{equation}

\subsection{Training and Inference}\label{sec:trainingandinference}
The total loss function for the ReIDMamba model, incorporating the proposed ranking-aware triplet regularization loss $\mathcal{L}_{ratr}$, is defined as follows:
\begin{equation}\label{eq:totalloss}
  \mathcal{L} =\frac{1}{G}\sum_{g=1}^G{\left( \mathcal{L} _{id}(\mathbf{f}^{(g)})+\mathcal{L} _{tri}(\mathbf{f}^{(g)}) \right)}+\rho \mathcal{L} _{ratr},
\end{equation}
where $\rho$ is a hyperparameter that balances the contribution of the ranking-aware triplet regularization loss. For inference, all L2-normalized features $\{\mathbf{f}^{(g)} \mid g=0,1,\cdots,G-1\}$ are concatenated to form the final representation.

\begin{table*}[!t]
  \caption{Comparison with the state-of-the-art models on MSMT17, Market-1501, DukeMTMC-reID, CUHK03 and Occluded-Duke datasets. Best results for previous methods in each part are highlighted in \underline{underline} and best of our methods in each part are highlighted in \textbf{bold}. The notation $\uparrow 384$ means that the input size is increased to $384 \times 128$, while $\uparrow 384^+$ indicates that the input size is increased to $384 \times 256$, and for the others, the input size remains at $256 \times 128$. The $^*$ denotes that the backbone uses a sliding window setting. The $^\dagger$ denotes methods that rely on extra clues from human pose and parsing models. The $^{\ddagger}$ denotes methods with occlusion instance augmentation.}
  \label{tab:sotas}
  \centering
  \begin{tabular}{@{}lcccccccccccccccccc@{}}
  \toprule
  \multicolumn{1}{c}{\multirow{3}{*}{Method}}                & \multicolumn{2}{c}{\multirow{2}{*}{MSMT17}} &  & \multicolumn{2}{c}{\multirow{2}{*}{Market1501}} &  & \multicolumn{2}{c}{\multirow{2}{*}{DukeMTMC-reID}} &  & \multicolumn{5}{c}{CUHK03}                                                   &  & \multicolumn{2}{c}{\multirow{2}{*}{Occluded-Duke}} &  \\ \cmidrule(lr){11-15}
  \multicolumn{1}{c}{}                                       & \multicolumn{2}{c}{}                        &  & \multicolumn{2}{c}{}                            &  & \multicolumn{2}{c}{}                               &  & \multicolumn{2}{c}{Labeled}         &  & \multicolumn{2}{c}{Detected}        &  & \multicolumn{2}{c}{}                               &  \\ \cmidrule(lr){2-3} \cmidrule(lr){5-6} \cmidrule(lr){8-9} \cmidrule(lr){11-12} \cmidrule(lr){14-15} \cmidrule(lr){17-18}
  \multicolumn{1}{c}{}                                       & mAP                  & R@1                  &  & mAP                    & R@1                    &  & mAP                      & R@1                     &  & mAP              & R@1              &  & mAP              & R@1              &  & mAP                      & R@1                     &  \\ \midrule \midrule
  \multicolumn{19}{l}{\textit{CNN-based methods}}                                                                                                                                                                                                                                                                                                                    \\ \midrule \midrule
  RGA-SC\cite{DBLP:conf/cvpr/ZhangLZJ020}                    & 57.5                 & 80.3                 &  & 88.4                   & \underline{96.1}       &  & -                        & -                       &  & \underline{77.4} & \underline{81.1} &  & \underline{74.5} & \underline{79.6} &  & -                        & -                       &  \\
  MSINet\cite{DBLP:conf/cvpr/Gu00C0FZ0023}                   & 59.6                 & 81.0                 &  & \underline{89.6}       & 95.3                   &  & -                        & -                       &  & -                & -                &  & -                & -                &  & -                        & -                       &  \\
  RFCNet$^\dagger$\cite{DBLP:journals/pami/HouMCGSC22}       & 60.2                 & 82.0                 &  & 89.2                   & 95.2                   &  & \underline{80.7}         & \underline{90.7}        &  & -                & -                &  & 78.0             & 81.1             &  & \underline{54.5}         & \underline{63.9}        &  \\
  MGN$\uparrow 384$\cite{DBLP:conf/mm/WangYCLZ18}            & 52.1                 & 76.9                 &  & 86.9                   & 95.7                   &  & 78.4                     & 88.7                    &  & 67.4             & 68.0             &  & 66.0             & 66.8             &  & -                        & -                       &  \\
  ABDNet$\uparrow 384$\cite{DBLP:conf/iccv/ChenDXYCYRW19}    & \underline{60.8}     & \underline{82.3}     &  & 88.3                   & 95.6                   &  & 78.6                     & 89.0                    &  & -                & -                &  & -                & -                &  & -                        & -                       &  \\ \midrule \midrule
  \multicolumn{19}{l}{\textit{Transformer-based methods}}                                                                                                                                                                                                                                                                                                            \\ \midrule \midrule
  TransReID\cite{DBLP:conf/iccv/He0WW0021}                   & 64.9                 & 83.3                 &  & 88.2                   & 95.0                   &  & 80.6                     & 89.6                    &  & -                & -                &  & -                & -                &  & 55.7                     & 64.2                    &  \\
  PFD$^{\dagger}$\cite{DBLP:conf/aaai/WangLS0S22}            & 65.1                 & 82.7                 &  & \underline{89.6}       & 95.5                   &  & \underline{82.2}         & \underline{90.6}        &  & -                & -                &  & -                & -                &  & 60.1                     & 67.7                    &  \\
  FCFormer$^{\ddagger}$\cite{DBLP:journals/tmm/WangLLLBGL24} & -                    & -                    &  & 86.8                   & 95.0                   &  & 78.8                     & 89.7                    &  & -                & -                &  & -                & -                &  & \underline{60.9}         & \underline{71.3}        &  \\
  TransReID$\uparrow 384$\cite{DBLP:conf/iccv/He0WW0021}     & \underline{66.6}     & \underline{84.6}     &  & 88.8                   & 95.0                   &  & 81.8                     & 90.4                    &  & -                & -                &  & -                & -                &  & -                        & -                       &  \\
  AAformer$\uparrow 384^+$\cite{zhu2023aaformer}             & 63.2                 & 83.6                 &  & 87.7                   & \underline{95.4}       &  & 80.0                     & 90.1                    &  & \underline{77.8} & \underline{79.9} &  & \underline{74.8} & \underline{77.6} &  & 58.2                     & 67.0                    &  \\ \midrule
  TransReID$^*$\cite{DBLP:conf/iccv/He0WW0021}               & 67.4                 & 85.3                 &  & 88.9                   & 95.2                   &  & 82.0                     & 90.7                    &  & -                & -                &  & -                & -                &  & 59.2                     & 66.4                    &  \\
  PFD$^{\dagger *}$\cite{DBLP:conf/aaai/WangLS0S22}          & 64.4                 & 83.8                 &  & 89.7                   & 95.5                   &  & \underline{83.2}         & \underline{91.2}        &  & -                & -                &  & -                & -                &  & \underline{61.8}         & \underline{69.5}        &  \\
  PHA$^*$\cite{DBLP:conf/cvpr/ZhangZZ0P23}                   & 68.9                 & 86.1                 &  & \underline{90.2}       & \underline{96.1}       &  & -                        & -                       &  & \underline{83.0} & \underline{84.5} &  & \underline{80.3} & \underline{83.2} &  & -                        & -                       &  \\
  TransReID$^*\uparrow 384$\cite{DBLP:conf/iccv/He0WW0021}   & \underline{69.4}     & \underline{86.2}     &  & 89.5                   & 95.2                   &  & 82.6                     & 90.7                    &  & -                & -                &  & -                & -                &  & -                        & -                       &  \\ \midrule \midrule
  \multicolumn{19}{l}{\textit{Mamba-based methods}}                                                                                                                                                                                                                                                                                                                  \\ \midrule \midrule
  ReIDMamba(ours)                                            & 66.4                 & 83.9                 &  & 89.1                   & 95.3                   &  & 82.1                     & 90.1                    &  & 82.3             & 84.3             &  & 78.7             & 81.5             &  & 56.3                     & 64.5                    &  \\
  ReIDMamba$\uparrow 384$(ours)                              & \textbf{68.6}        & \textbf{85.7}        &  & \textbf{89.6}          & \textbf{95.4}          &  & \textbf{82.9}            & \textbf{91.0}           &  & \textbf{82.5}    & \textbf{84.7}    &  & \textbf{80.2}    & \textbf{82.8}    &  & \textbf{58.6}                     & \textbf{67.0}                    &  \\ \midrule
  ReIDMamba$^*$(ours)                                        & 69.4                 & 86.4                 &  & 90.2                   & 95.5                   &  & 83.6                     & \textbf{91.5}           &  & 83.3             & 85.2             &  & 80.3    & 82.7             &  & 59.7                     & 67.4                    &  \\
  ReIDMamba$^*\uparrow 384$(ours)                            & \textbf{70.6}        & \textbf{87.0}        &  & \textbf{90.4}          & \textbf{95.7}                   &  & \textbf{83.9}            & \textbf{91.5}           &  & \textbf{83.4}    & \textbf{85.9}    &  & \textbf{80.6}             & \textbf{82.9}    &  & \textbf{60.8}                     & \textbf{68.3}               &  \\ \bottomrule
  \end{tabular}
\end{table*}

\section{Experiments}
We mainly evaluate the performance of ReIDMamba on five commonly used person ReID datasets like TransReID\cite{DBLP:conf/iccv/He0WW0021}, see Sec.\ref{sec:comparison}. The ablation studies of the Mamba-based baseline and ReIDMamba are presented in Sec.\ref{sec:mbb} and Sec.\ref{sec:reidmamba}, respectively.

\subsection{Datasets and Implementation Details}
\textbf{Dataset.} We evaluate our proposed method on five person ReID datasets, MSMT17\cite{DBLP:conf/cvpr/WeiZ0018}, Market-1501\cite{DBLP:conf/iccv/ZhengSTWWT15}, DukeMTMC-reID\cite{DBLP:conf/eccv/RistaniSZCT16}, CUHK03\cite{DBLP:conf/cvpr/LiZXW14} and Occluded-Duke\cite{DBLP:conf/iccv/MiaoWLD019}. The usage of these datasets is consistent with TransReID.

\textbf{Evaluation Protocols.} We follow the common practices and use the cumulative matching characteristics (CMC) at Rank-1 (R@1) and mean average precision (mAP) to evaluate the performance.

\textbf{Implementation.} We employ the Mamba$^\circledR$-S/16\cite{DBLP:journals/corr/abs-2405-14858} pre-trained on ImageNet\cite{DBLP:conf/cvpr/DengDSLL009} as the backbone network, where $D=384$. Unless otherwise specified, all person images are resized to $256\times 128$. The training images are augmented with random horizontal flipping, padding, random cropping and random erasing\cite{DBLP:conf/aaai/Zhong0KL020}. The batch size is set to 64 with $P=16, K=4$. The SGD optimizer is employed with a momentum of 0.9. The learning rate is initialized at 0.008 and employs a cosine learning rate decay strategy, which lasts for a total of 160 epochs. During the first 5 epochs, the learning rate undergoes a linear warm-up from 8e-5. The test-time flipping is utilized during testing, and the cosine similarity is utilized as the measurement.
Unless otherwise specified, we set $M=12, r=4, G=3, \rho=1, \tau=0.1$ for all experiments. All the experiments are performed with one NVIDIA GeForce RTX 4090 using the PyTorch with FP16 training.
The training time on the Market1501 dataset is approximately 2.67 hours.

\subsection{Comparison with State-of-the-Art Methods}\label{sec:comparison}
To comprehensively evaluate the performance of ReIDMamba, we compared it against previously reported state-of-the-art methods on person ReID datasets, as shown in Table \ref{tab:sotas}. Given that TransReID employs a small sliding window strategy in its patchify process, which is broadly adopted in later works, we present results under both settings. Additionally, we present results under the two common resolutions of $256 \times 128$ and $384 \times 128$. 
From the results in Table \ref{tab:sotas}, it can be observed that regardless of the resolution used and whether a small sliding window strategy is employed, ReIDMamba achieves a comparative performance with previous methods in handling the four holistic person ReID datasets MSMT17, Market1501, DukeMTMC-reID and CUHK03. Particularly on the MSMT17 dataset, ReIDMamba$^*\uparrow 384$ achieves an mAP of 70.6\%, surpassing the best performance of CNN-based and Transformer-based methods by 1.2\%. It is important to emphasize that, compared to methods such as PDF$^\dagger$\cite{DBLP:conf/aaai/WangLS0S22}, which introduce additional cues, FCFormer$^\ddagger$\cite{DBLP:journals/tmm/WangLLLBGL24}, which employs occlusion instance augmentation, and PHA\cite{DBLP:conf/cvpr/ZhangZZ0P23}, which uses carefully designed patch-wise high-frequency augmentation, our proposed ReIDMamba achieves better performance without these complex enhancements. This once again highlights the effectiveness of our model in the person ReID task.

In the Occluded ReID task, ReIDMamba outperforms TransReID, confirming the superiority of the Mamba structure in person ReID tasks. However, compared to methods such as PDF$^\dagger$\cite{DBLP:conf/aaai/WangLS0S22}, which introduce additional cues, and FCFormer$^\ddagger$\cite{DBLP:journals/tmm/WangLLLBGL24}, which employs occlusion instance augmentation, ReIDMamba, lacking these techniques, falls slightly behind in the Occluded ReID task. The main reason lies in the unique challenges posed by the occluded person ReID task, where the presence of occlusions can significantly affect the feature extraction process. In the absence of extra cues, it becomes more challenging to extract discriminative features in the Occluded ReID task. Despite this, these methods are also orthogonal to ReIDMamba. Therefore, it is entirely possible to integrate these techniques to further enhance ReIDMamba's performance on the Occluded-Duke dataset.

\begin{table*}[!t]
  \begin{minipage}[c]{0.65\textwidth}
    \centering
    \caption{Ablation study of training settings on MSMT17 and DukeMTMC-reID. The abbreviations BB, OPT, PE, SP, STL and LS denote Backbone, Optimizer, Position Embedding, Stochastic Depth, Soft Triplet Loss, Label Smoothing, respectively. Second-best results are highlighted in \underline{underline} and best results are highlighted in \textbf{bold}.}
    \label{tab:ablations1}
    \scalebox{0.95}{\begin{tabular}{@{}c|cccccc|cc|cc@{}}
      \toprule
      \multirow{2}{*}{Method}                                                          & \multirow{2}{*}{BB}                      & \multirow{2}{*}{OPT}       & \multirow{2}{*}{PE}                & \multirow{2}{*}{SP}                & \multirow{2}{*}{STL}               & \multirow{2}{*}{LS}                & \multicolumn{2}{c|}{MSMT17}   & \multicolumn{2}{c}{DukeMTMC-reID} \\
                                                                                       &                                          &                            &                                    &                                    &                                    &                                    & mAP           & R@1           & mAP             & R@1             \\ \midrule\midrule
      \multirow{2}{*}{Backbone}                                                        & Mamba$^\circledR$                        & SGD                        & \usym{2713}                          & \usym{2713}                          & \usym{2717}                          & \usym{2713}                          & \textbf{62.3} & \textbf{82.1} & \textbf{79.9}   & \textbf{89.1}   \\ %\midrule
                                                                                       & Vim                                      & \textcolor{lightgray}{SGD} & \textcolor{lightgray}{\usym{2713}} & \textcolor{lightgray}{\usym{2713}} & \textcolor{lightgray}{\usym{2717}} & \textcolor{lightgray}{\usym{2713}} & 59.6          & 80.3          & 78.4            & 87.5            \\ \midrule
      \multirow{2}{*}{Optimizer}                                                       & \textcolor{lightgray}{Mamba$^\circledR$} & Adam                       & \textcolor{lightgray}{\usym{2713}} & \textcolor{lightgray}{\usym{2713}} & \textcolor{lightgray}{\usym{2717}} & \textcolor{lightgray}{\usym{2713}} & 61.7          & 81.7          & 78.8            & 88.6            \\
                                                                                       & \textcolor{lightgray}{Mamba$^\circledR$} & AdamW                      & \textcolor{lightgray}{\usym{2713}} & \textcolor{lightgray}{\usym{2713}} & \textcolor{lightgray}{\usym{2717}} & \textcolor{lightgray}{\usym{2713}} & 61.6          & 81.6          & 78.6            & 88.5            \\ \midrule
      \multirow{2}{*}{\begin{tabular}[c]{@{}c@{}}Network\\ Configuration\end{tabular}} & \textcolor{lightgray}{Mamba$^\circledR$} & \textcolor{lightgray}{SGD} & \usym{2717}                          & \textcolor{lightgray}{\usym{2713}} & \textcolor{lightgray}{\usym{2717}} & \textcolor{lightgray}{\usym{2713}} & 61.9          & 81.8          & 79.1            & 88.7            \\
                                                                                       & \textcolor{lightgray}{Mamba$^\circledR$} & \textcolor{lightgray}{SGD} & \textcolor{lightgray}{\usym{2713}} & \usym{2717}                          & \textcolor{lightgray}{\usym{2717}} & \textcolor{lightgray}{\usym{2713}} & 61.8          & 81.7          & 79.5            & 88.9            \\ \midrule
      \multirow{2}{*}{Loss Function}                                                   & \textcolor{lightgray}{Mamba$^\circledR$} & \textcolor{lightgray}{SGD} & \textcolor{lightgray}{\usym{2713}} & \textcolor{lightgray}{\usym{2713}} & \usym{2713}                          & \textcolor{lightgray}{\usym{2713}} & 62.0          & \underline{82.0}    & \underline{79.6}      & \underline{89.0}      \\
                                                                                       & \textcolor{lightgray}{Mamba$^\circledR$} & \textcolor{lightgray}{SGD} & \textcolor{lightgray}{\usym{2713}} & \textcolor{lightgray}{\usym{2713}} & \textcolor{lightgray}{\usym{2717}} & \usym{2717}                          & \underline{62.1}    & 81.9          & 79.4            & 88.9            \\ \bottomrule
      \end{tabular}}
  \end{minipage}% \hspace{1.0cm}
  \begin{minipage}[c]{0.35\textwidth}
    \centering
    \caption{Ablation study of architecture configurations on MSMT17 and DukeMTMC-reID. Second-best results are highlighted in \underline{underline} and best results are highlighted in \textbf{bold}.}
    \label{tab:ablations2}
    \scalebox{0.85}{
      \begin{tabular}{@{}ccc|cc|cc@{}}
        \toprule
        \multicolumn{3}{c|}{Cfgs.}                  & \multicolumn{2}{c|}{MSMT17}   & \multicolumn{2}{c}{DukeMTMC-reID} \\
        $M$                     & $r$ & Ind. Tr.    & mAP           & R@1           & mAP            & R@1              \\ \midrule\midrule
        1                       & 1   & -           & 59.7          & 80.4          & 78.6           & 87.8             \\
        2                       & 1   & \usym{2713} & 60.3          & 80.8          & 78.9           & 88.1             \\
        4                       & 1   & \usym{2713} & 60.4          & 80.7          & 79.0           & 88.2             \\ \midrule
        2                       & 1   & \usym{2717} & 61.5          & 81.6          & 79.3           & 88.4             \\
        4                       & 1   & \usym{2717} & 61.7          & 81.8          & 79.5           & 88.6             \\
        6                       & 1   & \usym{2717} & 61.8          & 81.7          & 79.4           & 88.5             \\
        4                       & 2   & \usym{2717} & 61.6          & 81.5          & 79.5           & 88.7             \\
        8                       & 2   & \usym{2717} & 62.0          & 81.8          & 79.7           & 89.0             \\
        12                      & 2   & \usym{2717} & \textbf{62.3} & \textbf{82.1} & \underline{79.9}     & \underline{89.1}       \\
        \rowcolor{lightgray} 12 & 4   & \usym{2717} & \underline{62.2}    & \textbf{82.1} & \textbf{80.0}    & \textbf{89.2}    \\
        12                      & 6   & \usym{2717} & 61.9          & \underline{81.9}    & 79.8           & 88.9             \\ \bottomrule
        \end{tabular}}
  \end{minipage}
  
\end{table*}

\subsection{Ablation Study of Mamba-based Strong Baseline}\label{sec:mbb}
In this section, we primarily conduct a detailed analysis of the Mamba-based strong baseline. We start by examining the training settings of the Mamba-based models and then delve into the architecture configurations within the Mamba-based strong baseline.

\textbf{Training settings.} We conducted a detailed analysis of the settings for training the Mamba-based models. Ablation studies are presented in Table \ref{tab:ablations1}, showing the performance on MSMT17 and DukeMTMC-reID with various training settings. Firstly, our model initialization mainly compared two pure Mamba-based models: Mamba$^\circledR$-S/16 and Vim-S/16. Without changing the backbone structure, Mamba$^\circledR$-S/16 outperformed the other model. Next, we compared three different optimizers: SGD, Adam, and AdamW. Consistent with the results obtained from TransReID, the SGD optimizer performed best on the Mamba-based model. Position embeddings and stochastic depth remain two effective techniques. However, while TransReID sets the stochastic depth probability to 0.1, a setting of 0.3 in the Mamba-based strong baseline yielded better performance. Regarding the loss function, Mamba-based strong baseline also shows inconsistency with TransReID, using soft triplet loss and not employing label smoothing both lead to a decrease in performance accuracy. Setting the triplet loss margin to 1.2 and using label smoothing can bring about better performance. This suggests that the Mamba network, compared to the Transformer network, is more susceptible to overfitting and therefore benefits from a higher stochastic depth probability and the use of label smoothing. The first row in Table \ref{tab:ablations1} shows the default training settings for the Mamba-based strong baseline, which are also used for ReIDMamba.

\textbf{Architecture configurations.} We conducted a detailed analysis of the number of class tokens $M$ and the reduction factor $r$ in the Mamba-based strong baseline, as shown in Eq.\ref{eq:ss3}. Additionally, we explored the approach of training each class token independently (Ind. Tr. as shown in Table \ref{tab:ablations2}), which is similar to TransReID. Ablation studies are presented in Table \ref{tab:ablations2}, showing the performance on MSMT17 and DukeMTMC-reID with various architecture configurations. When $M=r=1$, the architecture configuration is consistent with Vim. Although the initialization parameters are from Mamba$^\circledR$, leading to different final results from Table \ref{tab:ablations1}, the overall performance remains essentially the same. The configuration of $M=12, r=2$ is the default architecture setting for Mamba$^\circledR$-S/16, aligning with the 'Baseline' results in Table \ref{tab:ablations1}. Firstly, it can be observed from Table \ref{tab:ablations2} that training class tokens independently leads to a significant decrease in performance. Concatenating the class tokens before joint training proves to be an effective strategy. Moreover, when $M/r$ is held constant, an increase in $M$ initially improves performance but then declines. An appropriate number of class tokens aids in extracting fine-grained features, thereby enhancing performance. When $r$ is fixed, increasing $M$ consistently improves performance. Considering the feature dimension $MD/r$ and the upcoming introduction of a multi-branch architecture in ReIDMamba, we set $M$ to the default configuration of 12 in Mamba\(^\circledR\)-S/16. At this point, the configurations of $M=12, r=2$ and $M=12, r=4$ exhibit comparable performance. It's evident that the $M=12, r=4$ configuration strikes a good balance between accuracy and dimensionality. Therefore, this configuration has been chosen for the Mamba-based strong baseline and ReIDMamba, and \textbf{we refer to it as the Baseline in subsequent studies} (highlighted in light gray as shown in Table \ref{tab:ablations2}). The ordered nature of patches in Mamba-based models, compared to Transformer-based models, means that relying solely on one class token is insufficient for the fine-grained metric learning task of person ReID. Capturing detailed information about individuals is challenging, making it crucial to distribute a certain number of class tokens uniformly within Mamba-based models.

\begin{table}[!t]
  \centering
  \caption{Ablation study of each component in ReIDMamba on the MSMT17 and DukeMTMC-reID. The abbreviations Baseline, $\mathcal{M}$, $\mathcal{R}$ denote the Mamba-based strong baseline, MGFE, and RATR, respectively. The notation $\mathcal{M}(M, r, G)$ represents the configuration settings within ReIDMamba, where $\mathcal{R}$-intra and $\mathcal{R}$-inter denote the intra-class diversity constraint loss and inter-class diversity constraint loss, respectively. Dim. denote the dimension of the final feature. Second-best results are highlighted in \underline{underline} and best results are highlighted in \textbf{bold}.}
\label{tab:ablation3}
\begin{tabular}{@{}l|c|cc|cc@{}}
\toprule
\multicolumn{1}{c|}{\multirow{2}{*}{Settings}} & \multirow{2}{*}{Dim.} & \multicolumn{2}{c|}{MSMT17}   & \multicolumn{2}{c}{DukeMTMC-reID} \\
\multicolumn{1}{c|}{}                          &                       & mAP           & R@1           & mAP             & R@1             \\ \midrule\midrule
Baseline                                       & 1152                  & 62.2          & 82.1          & 80.0            & 89.2            \\
+$\mathcal{M}$(12,6,3)                           & 2304                  & 65.1          & 83.3         & 80.7            & 89.5            \\
+$\mathcal{M}$(12,4,3)                           & 3456                  & 65.5          & 83.5          & 81.1            & 89.7            \\
+$\mathcal{M}$(12,2,3)                           & 6912                  & 65.4          & 83.6          & 80.8            & 89.5            \\ \midrule
+$\mathcal{M}$(12,4,2)                           & 2304                  & 65.0          & 83.3          & 80.6            & 89.4            \\
+$\mathcal{M}$(12,4,4)                           & 4608                  & 65.6          & 83.6          & 81.1            & 89.8            \\ \midrule
+$\mathcal{M}$(12,4,3)+$\mathcal{R}$-intra         & 3456                  & \underline{66.1}    & \underline{83.7}    & \underline{81.7}      & \underline{89.9}      \\
+$\mathcal{M}$(12,4,3)+$\mathcal{R}$-inter         & 3456                  & 65.9          & 83.6          & 81.5            & 89.8            \\
+$\mathcal{M}$(12,4,3)+$\mathcal{R}$               & 3456                  & \textbf{66.4} & \textbf{83.9} & \textbf{82.1}   & \textbf{90.1}   \\ \bottomrule
\end{tabular}

\end{table}

\subsection{Ablation Study of ReIDMamba}\label{sec:reidmamba}
\textbf{Effectiveness of each component.} Using the settings of $M=12, r=4$ from Table \ref{tab:ablations2} as the baseline, we first considered the scenario with $G=3$ and compared the three settings with $r=2, 4, 6$. It can be observed that the addition of the MGFE component to the baseline model resulted in performance improvements across all three settings. Subsequently, under the setting of $M=12, r=4$, we compared the performance with $G$ set to 2, 3, and 4. The results indicate that performance is nearly proportional to the size of $G$. However, when $G=4$, the performance is nearly saturated. This setting, however, results in a high-dimensional feature space of 4608 dimensions. As a result, the final setting for ReIDMamba in MGFE is determined to be $M=12, r=4, G=3$. Lastly, in analyzing the effectiveness of each component in RATR, the results indicate that the intra-class diversity constraint loss provides greater benefits compared to the inter-class diversity constraint loss. This suggests that in ReID tasks, exploring diversity among positive samples yields more effective results than among negative samples. However, both types of diversity constraint losses are indispensable, and using them together yields the best performance. It is important to note that, compared to TransReID's 3840-dimensional feature, our final setting for ReIDMamba has a feature dimension of 3456. Despite a decrease of 384 dimensions, the performance improvement of ReIDMamba is still significant, as shown in Table \ref{tab:sotas}. This demonstrates that ReIDMamba is able to achieve substantial performance gains with a relatively lower-dimensional feature space, highlighting the efficiency and effectiveness of the proposed method.

\begin{figure}[!t]
  \centering
  \includegraphics[width=0.5\textwidth]{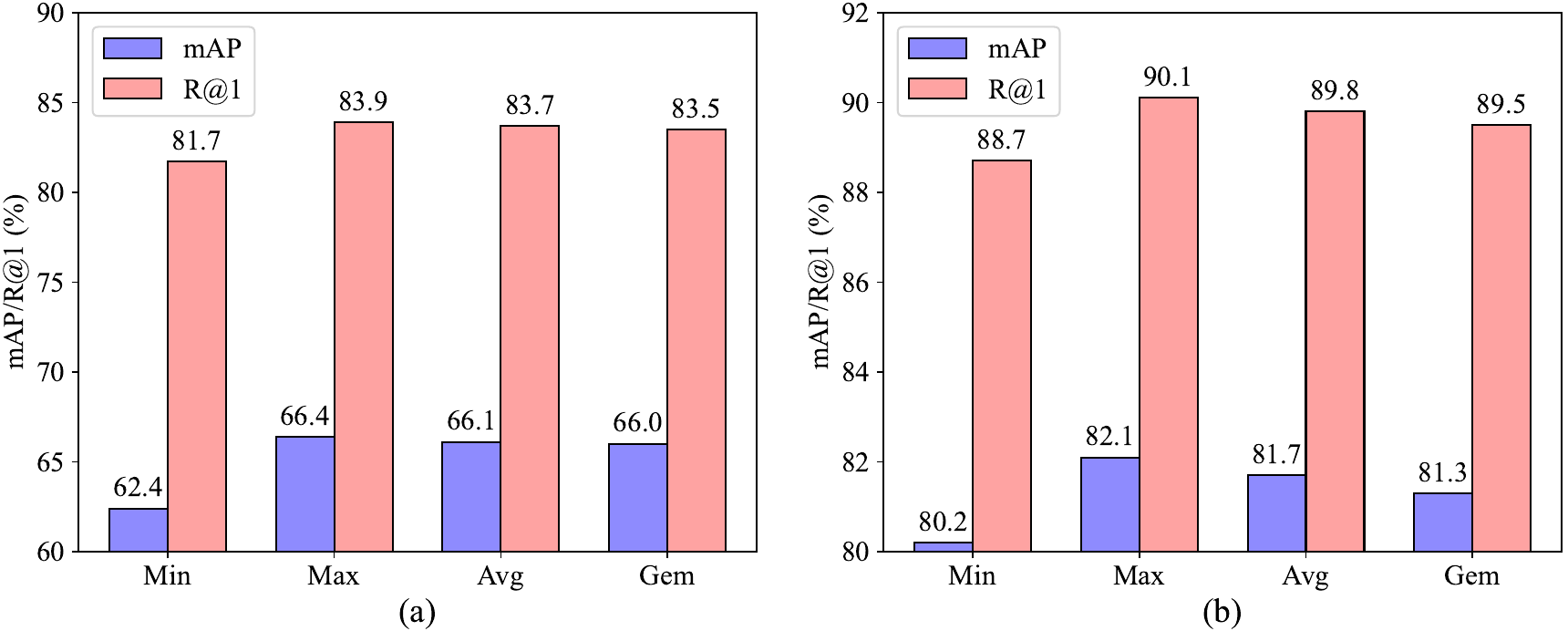}
  \caption{Comparison of different tokens fusion operations on (a) MSMT17 and (b) DukeMTMC-reID datasets.}
  \label{fig:5}
\end{figure}

\begin{figure}[!t]
  \centering
  
  \includegraphics[width=0.5\textwidth]{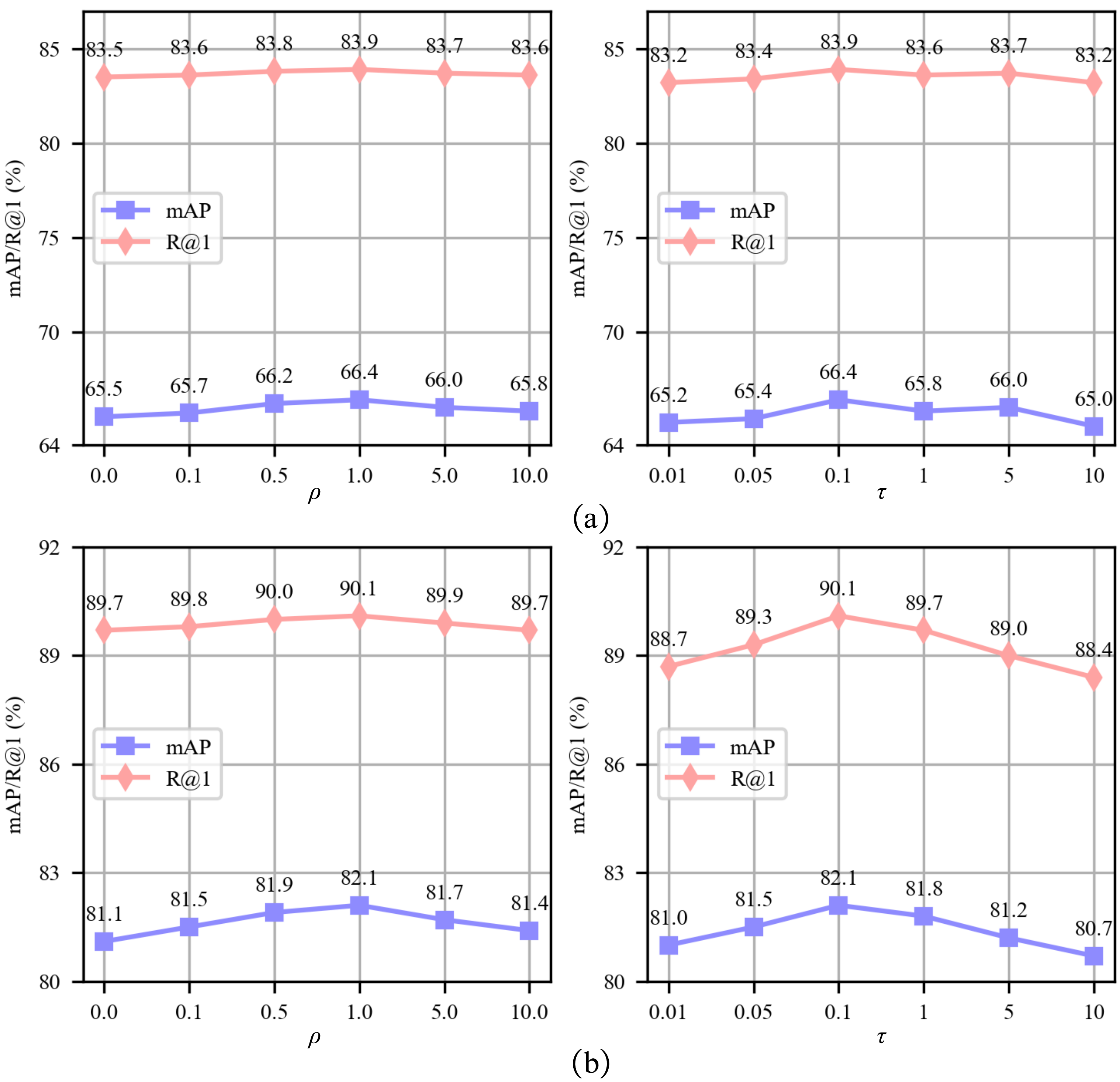}
  \caption{Impact of hyperparameters in RATR on (a) MSMT17 and (b) DukeMTMC-reID datasets.}
  \label{fig:6}
  
\end{figure}

\textbf{Comparison of different tokens fusion.} We conducted an in-depth comparative analysis of tokens fusion operations, focusing on minimum, maximum, average, and generalized-mean (Gem) pooling \cite{DBLP:journals/pami/YeSLXSH22}. Gem, which includes a learnable parameter, is designed to operate somewhere between max pooling and avg pooling. The results shown in Fig. \ref{fig:5} indicate that the minimum operation performed the worst, likely because this operation is easily dominated by the least informative class tokens, thereby causing a decrease in performance. Conversely, the maximum operation emerged as the best performer, possibly because it captures the most salient and discriminative information from the class tokens, ensuring that the most activated tokens contribute to the final representation. The Gem operation, despite its flexibility due to the learnable parameter, did not outperform the other operations, suggesting that additional complexity may not always lead to improved performance and could potentially result in overfitting or increased sensitivity to parameter initialization. The average operation, a commonly used method, demonstrated moderate performance, indicating that while it provides an average representation of tokens, it might not emphasize the most informative tokens as effectively as the maximum operation.
Based on these findings, the maximum operation is adopted as the tokens fusion method in ReIDMamba.

\begin{figure*}[!t]
  \centering
  
  \includegraphics[width=1.0\textwidth]{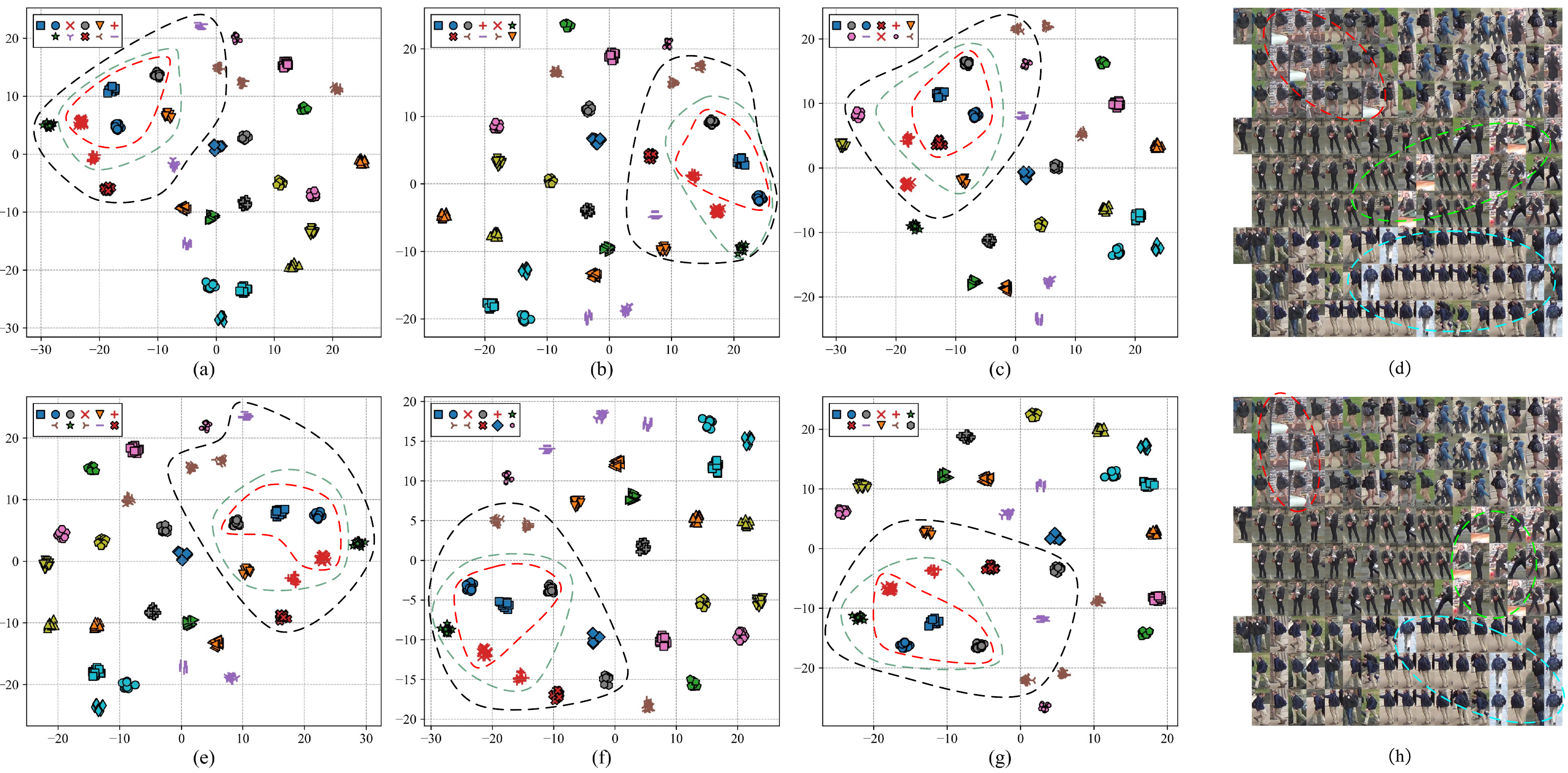}
  \caption{The visualizations of ReIDMamba and ReIDMamba without RATR. (a), (b), (c) The tSNE embeddings of the three branch features of ReIDMamba (In the top-left corner, the 10 most similar categories relative to the blue square are ranked from most to least similar. The red, green, and black dashed circles respectively enclose the areas of the top 3, 5, and 10 most similar categories.). (d) The intra-class similarity ranking of ReIDMamba's three branch features (The first image in the first row serves as the anchor image, followed by 17 columns of images sorted in descending order of similarity. The features of branch 1, 2, and 3 correspond to rows 1, 2, and 3, respectively.
  ). (e), (f), (g) The tSNE embeddings of the three branch features of ReIDMamba without RATR. (h) The intra-class similarity ranking of the three branch features of ReIDMamba without RATR.
  (Best viewed in color.)}
  \label{fig:7}
  
\end{figure*}

\begin{table}[!t]
  \centering
  \caption{Model comparison between TransReID$^*\uparrow 384$ and our ReIDMamba$^*\uparrow 384$. The throughput values are measured with a 4090 GPU and an Intel(R) Xeon(R) Platinum 8352V CPU. The memory overhead is measured with a batch size of 128 on a single GPU.}
  \label{tab:ablation4}
  \begin{tabular}{@{}l|cc@{}}
    \toprule
                        & TransReID$^*\uparrow 384$ & ReIDMamba$^*\uparrow 384$ \\ \midrule
    Param.(M)          & 92.98                     & 29.64\textcolor{green}{($\downarrow$ 63.34)}                     \\
    FLOPs (G)           & 30.66                     & 10.06\textcolor{green}{($\downarrow$ 20.6)}                     \\
    Throughput(image/s) & 489.27                    & 536.84\textcolor{green}{($\uparrow$ 47.57)}                    \\
    Mem.(G)             & 2.23                      & 1.74\textcolor{green}{($\downarrow$ 0.49)}                      \\ \bottomrule
    \end{tabular}
  
\end{table}

\begin{figure}[!t]
  \centering
  
  \includegraphics[width=0.5\textwidth]{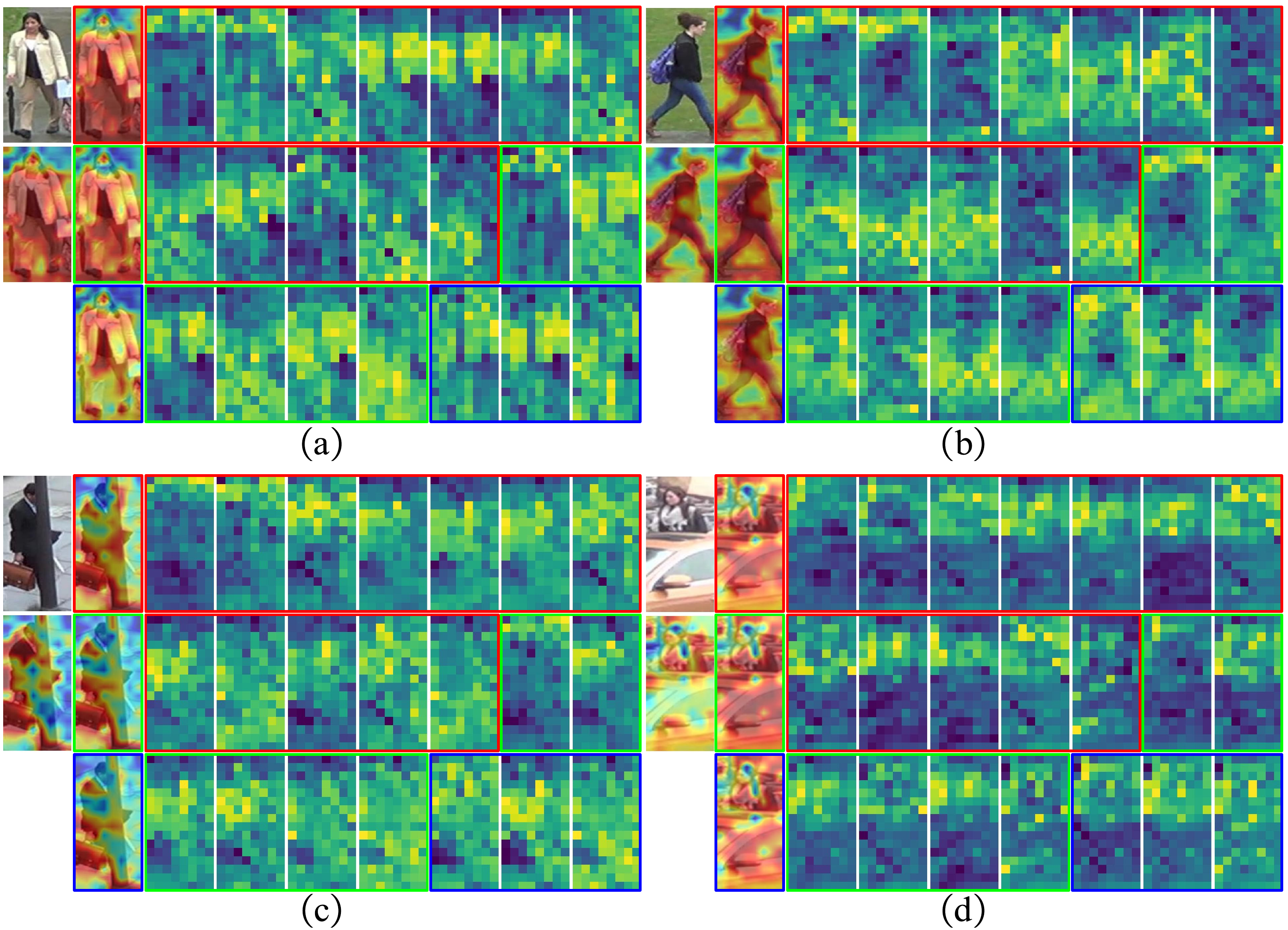}
  \caption{Feature maps for different class tokens and Grad-CAM\cite{DBLP:journals/ijcv/SelvarajuCDVPB20} for ReIDMamba. The first column displays the original images and the overall Grad-CAM visualization results. The second column shows the Grad-CAM visualization results for the three branches. Subsequently, the seven columns feature the visualization results of the feature maps for the first branch's 12 class tokens (enclosed in red boxes), the second branch's 6 class tokens (enclosed in green boxes), and the third branch's 3 class tokens (enclosed in blue boxes).
  }
  \label{fig:8}
  
\end{figure}

\textbf{Impact of hyperparameters in RATR.} The hyperparameters $\rho$ (see Eq. \ref{eq:totalloss}) and $\tau$ (see Eq. \ref{eq:dktau}) involved in RATR were analyzed, as shown in Fig. \ref{fig:6}. On both the MSMT17 and DukeMTMC-reID datasets, with $\tau$ fixed at 0.1, the performance initially increased and then decreased as $\rho$, the weight of the RATR loss function, gradually increased, peaking at $\rho = 1.0$. When analyzing the impact of the hyperparameter $\tau$, with $\rho$ fixed at 1.0, the performance on both datasets initially increased and then decreased, peaking at $\tau = 0.1$. Additionally, observing the overall curve changes, the hyperparameter $\rho$ exhibited greater robustness, indicating that the model's performance remained consistently good across a broad range of $\rho$ values as long as RATR was incorporated. The default settings for ReIDMamba are $\rho = 1.0$ and $\tau = 0.1$.

\textbf{Analysis of the model cost.} 
Following the protocol proposed in \cite{liu2024vmamba}, we compared the number of parameters, FLOPs, throughput, and GPU memory usage during the inference phase between TransReID$^*\uparrow 384$ and ReIDMamba$^*\uparrow 384$. The results from Table\ref{tab:ablation4} indicate that ReIDMamba$^*\uparrow 384$ has significant advantages over TransReID$^*\uparrow 384$ in terms of inference speed (10\% faster) and inference GPU consumption (20\% less). These advantages become more pronounced with higher resolution images, as shown in Fig.\ref{fig:1}. However, due to GPU memory limitations, we were unable to explore the performance of ReIDMamba at even higher resolutions. Nevertheless, ReIDMamba provides a lightweight approach for large-scale person search tasks, as referenced in \cite{DBLP:journals/tmm/YangWY24}.

\textbf{Visualization results.}
We compared the visualization results of ReIDMamba with and without RATR. From Fig.\ref{fig:7}, a qualitative analysis shows that the diversity of intra-class and inter-class rankings has been significantly improved after employing RATR. For example, the top 5 other classes ranked relative to the blue square by the three branches of ReIDMamba include 7 different classes, while the three branches of ReIDMamba without RATR include only 6 different classes. Similar results can also be observed in inter-class rankings. Subsequently, a quantitative analysis was conducted by calculating the KTau metric; without using RATR, the intra-class and inter-class KTau values are 0.8377 and 0.7731, respectively. However, after adopting RATR, the intra-class and inter-class KTau values are reduced to 0.6472 and 0.6883, respectively. Furthermore, the overall model performance has been further enhanced, as shown in Table \ref{tab:ablation3}, confirming the effectiveness of the RATR module.

We have also visualized the feature maps of the three branches in ReIDMamba and utilized Grad-CAM to visualize the regions of interest for each branch as well as the combination of all branches, as shown in Fig. \ref{fig:8}. From the Grad-CAM visualization results, it can be observed that regardless of the presence of occlusions, the model is able to focus on key areas of the persons and extract discriminative features. Looking at the feature maps for different class tokens, it is evident that each branch's class tokens play a role in mining features with multiple granularities, thereby further enhancing the robustness of the features.

\section{Conclusion}
In this paper, our ReIDMamba framework, built on a pure Mamba approach, has made significant strides in person re-identification by overcoming the scalability limitations of traditional Transformer-based methods. ReIDMamba introduces innovative techniques like the multi-granularity feature extractor (MGFE) and ranking-aware triplet regularization (RATR) to enhance feature robustness and discrimination. 

Looking forward, the pre-training of ReID models based on Mamba presents a promising avenue for future research. This approach could optimize the balance between accuracy and computational efficiency, leading to more advanced and scalable systems for real-world applications. ReIDMamba's success indicates the potential for Mamba-driven methods to revolutionize person re-identification and related fields.

\bibliography{refs.bib}
\bibliographystyle{IEEEtran}

\end{document}